\documentclass[letterpaper, 10 pt, conference]{ieeeconf} 

\usepackage{cite}
\usepackage{hyperref}
   \usepackage[pdftex]{graphicx}
   \graphicspath{{./img/}{../jpeg/}}

\usepackage{amsmath}

\usepackage{color,soul}

\title{A Dataset of Stationary, Fixed-wing Aircraft on a Collision Course for Vision-Based Sense and Avoid}

\author{J. Martin$^1$, J. Riseley$^1$ and J.J. Ford$^1$%
\thanks{$^1$J. Martin, J. Riseley, and  J. J. Ford and are with the School of Electrical Engineering and Robotics, Queensland University of Technology, 2 George St, Brisbane QLD, 4000 Australia. {\tt\small jasmin.martin@qut.edu.au,  j2.riseley@qut.edu.au, j2.ford@qut.edu.au}.}%
}

\begin{document}
\def\UrlFont{\em}

\maketitle
\thispagestyle{empty}
\pagestyle{empty}
\begin{abstract}
The emerging global market for  unmanned aerial vehicle (UAV) services is anticipated to reach USD 58.4 billion by 2026, spurring significant efforts to safely integrate routine UAV operations into the national airspace in a manner that they do not compromise the existing safety levels.
The commercial use of UAVs would be enhanced by an ability to sense and avoid potential mid-air collision threats however research in this field is hindered by the lack of available datasets as they are expensive and technically complex to capture.
In this paper we present a dataset for vision based aircraft detection. The dataset consists of 15  image sequences containing 55,521 images of a fixed-wing aircraft approaching a stationary, grounded camera. Ground truth labels and a performance benchmark are also provided. To our knowledge, this is the first public dataset for studying medium sized, fixed-wing aircraft on a collision course with the observer. The full dataset and ground truth labels are publicly available at \textit {\href{https://qcr.github.io/dataset/aircraft-collision-course/}{https://qcr.github.io/dataset/aircraft-collision-course/}}.
\end{abstract}

\section{Introduction}
The overall global market for unmanned aerial vehicle (UAV) services is estimated to be USD 27.4 billion in 2021 and is projected to reach USD 58.4 billion by 2026 with important use in commercial, government and law enforcement, and consumer applications \cite{Marketsandmarkets}.
This rapid growth has spurred research into safely integrating routine UAV operations into the national airspace in a manner that they do not compromise the existing safety levels \cite{clothier2015structuring}. One of the key risks that pose safety concerns is the risk of a collision with another aircraft in mid-air. 

Encounters between manned aircraft are currently avoided through multiple mechanisms. The first layers are regulation (rules that reduce the likelihood of encounters) and air traffic control (online traffic routing in radio communication with aircraft).
The final safety layer is what we consider in this paper and encompasses non-cooperative, potential mid-air collisions that have been missed or otherwise failed by the other layers. 
For manned aircraft with human pilots, this final layer involves the pilot visually seeing a potential mid-air collision and taking avoidance action.
For unmanned aircraft, there is the implied regulatory requirement that UAVs be capable of avoiding potential collision threats with a capability of autonomous sense and avoid (SAA) which meet or exceed the ability of a human pilot  \cite{james2018learning}.
Developing these SAA systems that are capable of matching (and exceeding) the  performance of human pilots is one of the key technical challenges hindering the routine, standard and flexible operation of UAVs in the national airspace \cite{clothier2015structuring}.

\begin{figure}
\begin{center}
\includegraphics[scale=0.8]{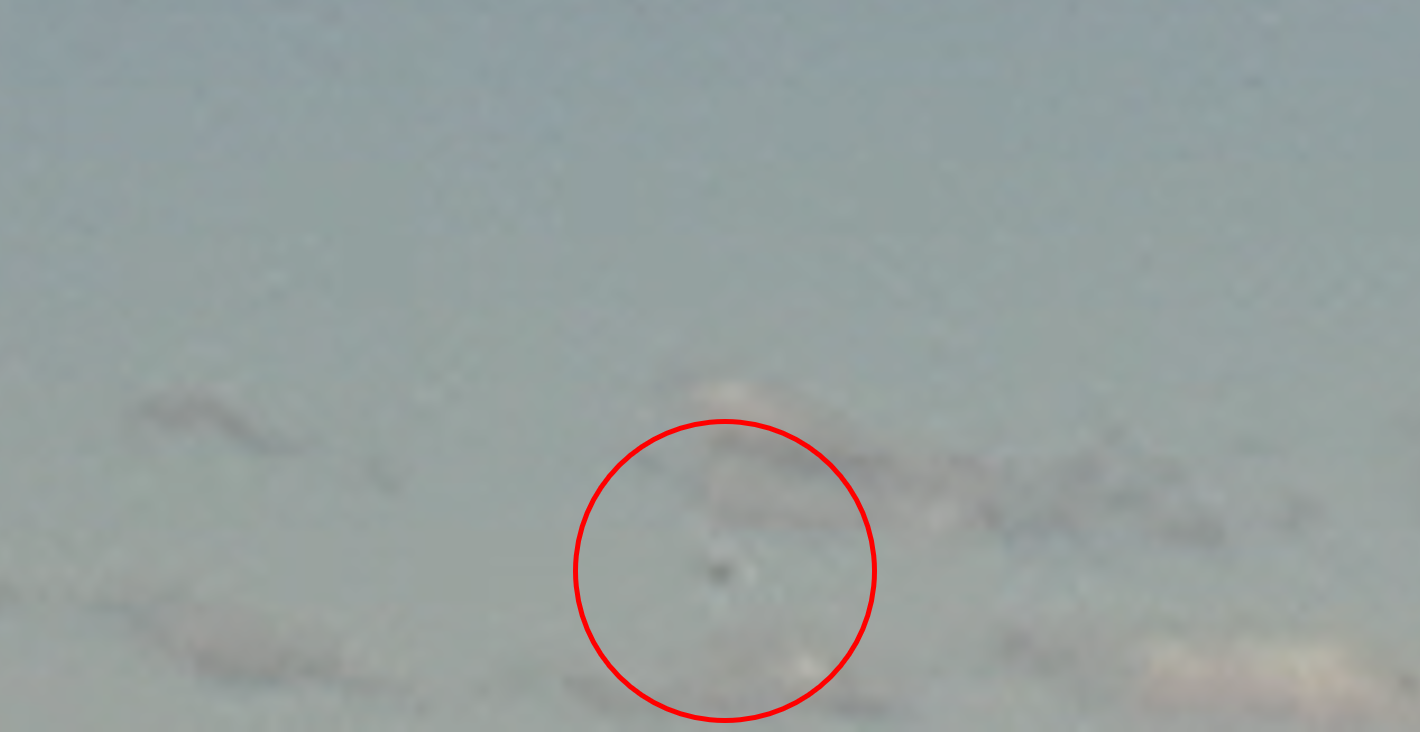}
 \caption{An illustration of an aircraft from our dataset (S2) at a detectable distance where it visually appears as a small of number locally dim pixels.}
\label{fig:exampleAircraft}
\end{center}
\end{figure}

With the falling size and cost of digital cameras, vision-based techniques are the preferred choice for sense and avoid in small to medium sized UAVs over other sensing approaches such as radar \cite{molloy2017detection}. 
Vision based SAA systems need to be able  detect aircraft at distances where the visually appear as a small number of locally dim pixels as shown in Figure \ref{fig:exampleAircraft}, so that there is time for avoidance manoeuvers.

\begin{figure}
\begin{center}
\includegraphics[scale=0.45]{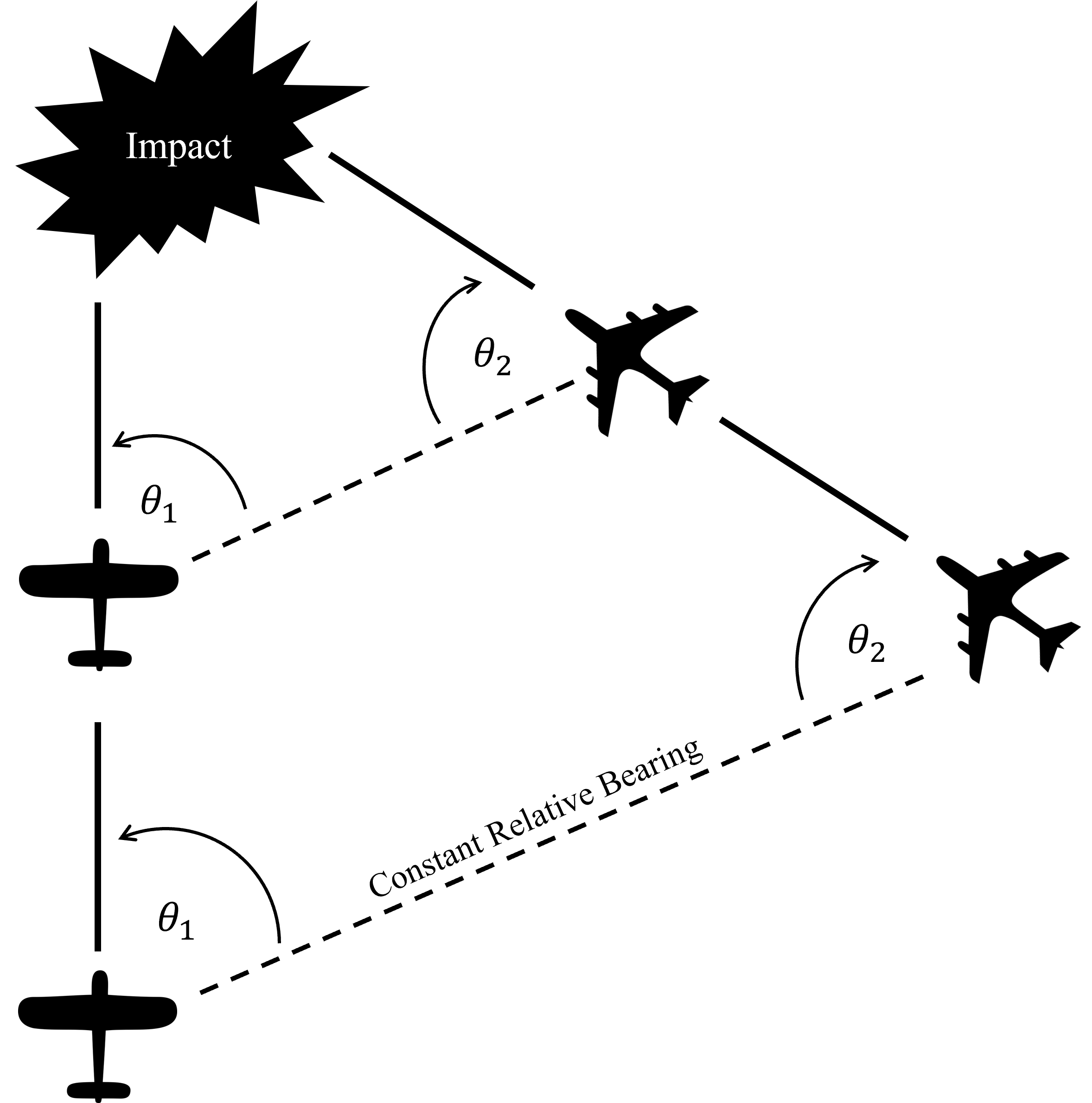}
 \caption{An illustration of an aircraft with constant relative bearing in a collision course scenario, adapted from \cite{laiStationaryTarget}.}
\label{fig:collision}
\end{center}
\end{figure}

A key challenge in vision based aircraft detection is how aircraft on a true collision appear in an image sequence. If two aircraft are on a true collision course and are converging on a common point of impact as seen in Figure \ref{fig:collision}, then the relative bearings of each aircraft from the other will remain constant up to the point of  collision \cite{laiStationaryTarget}. 
Hence, from each pilot’s point of view the other aircraft will exhibit no apparent motion and remain stationary with respect to the windscreen (zero relative bearing rate).
In our vision-based aircraft detection problem this results in aircraft on a  true collision course exhibiting very little (if any) motion in the image plane with respect to the background. Aside from the small size and stationary nature of the target, other challenges faced by vision based aircraft detection systems are: multiple simultaneous target aircraft in the image, changes in weather and lighting conditions where the target aircraft can exhibit very different visual feature, and  different backgrounds: the sky, ground clutter, and the interface between these two backgrounds. 

Another unique aspect of this detection problem is that the small size and low contrast at the desired detection ranges requires exploitation of temporal features of object on collision course. However, as highlighted in foundation work on diffusion of context and vanishing gradient, creating or learning detection of long temporal dependencies is difficult \cite{Bengio}.  %

Currently,  research in this field is  hindered by the lack of available datasets as, due to their nature, they are expensive and technically complex to capture. The only publicly available data to test on for vision based aircraft detection is \cite{carrio14} which is a simulated dataset of 24 encounters and \cite{Minwalla16} which only includes 2 encounters with fixed wing aircraft which are not on a collision course with the observer (and 2 rotorcraft).

In this paper we present a dataset for vision based aircraft detection for small to medium sized, fixed-wing aircraft. The dataset consists of 15 uncompressed, high-resolution image sequences of a fixed-wing aircraft approaching a stationary, grounded camera. Ground truth labels and range information are also provided. Our datatset includes  stationary target aircraft, multiple simultaneous target aircraft, different target appearance and difference cloud conditions. We also provide a benchmark performance from a state of the art vision based aircraft detection system presented in \cite{jamesCST}.

The rest of this paper is structured as follows. In Section \ref{sec:litrev} we discuss the related literature. In Section \ref{sec:dataCollection} we describe our data collection process. In Section \ref{sec:dataset} we present our dataset for vision based aircraft detection including the structure, ground truth labels and some commentary. In Section \ref{sec:benchmark} we present a benchmark study of a state of the art approach to vision based aircraft detection for small to medium fixed wing aircraft. In Section \ref{sec:conc} we provide concluding remarks.

\section{Related Work} \label{sec:litrev}
Numerous approaches to vision-based aircraft detection have been presented in the literature with the most effective methods utilising a multi-stage detection pipeline primarily consisting of image pre-processing and temporal filtering \cite{Carnie2006,Lai2013,Nussberger2014,molloy2017,jamesCST,Petridis2008}.

The goal of the  image pre-processing stage is to suppress background clutter and highlight small pixel sized aircraft. Some popular approaches that are proposed in the literature to accomplish this include morphology \cite{Carnie2006,Lai2013,Nussberger2014,jamesCST}, image frame differencing \cite{molloy2017,Nussberger2014}, and machine and deep learning \cite{Petridis2008,james2018learning, Dey2011, Sapkota2016, Opromolla19, Rozantsev2014, james2019below, Hussaini}. As one of the key outcomes of this stage is to separate the image background from potential aircraft targets it is essential that these methods are tested on data with true head on collision course encounters where the aircraft appears stationary with respect to the background.
Image frame differencing approaches would likely fail when confronted with a true collision course (stationary) aircraft, as would learnt approaches trained on moving targets. Our dataset captures this stationary characteristic of aircraft of a true collision course encounter.

Due to the low SNR of small pixel sized aircraft,  the temporal filtering stage is required to emphasize and extract features that possess aircraft-like dynamics. In \cite{Carnie2006}  a Viterbi-based filtering approach is proposed. The authors report detection ranges of around $6$km, however they do not report the false alarm rate. An extended Kalman filter is proposed in \cite{Nussberger2014}; a ``valid track'' of the aircraft (where the aircraft is consistently detected) is declared at an average detection range of $1747$m with an average of $4$ false alarms over their tested image sequences. In \cite{Lai2013} hidden Markov model (HMM) filtering is used for detection and the authors report detection ranges of at least $1540$m with no false alarms.

The literature around vision based aircraft detection are typically evaluated on either simulated data or a small number of real aircraft targets which are generally not available for other researchers to test on. This makes comparative studies challenging and does not provide sufficient insight into the performance on true collision course encounters. 
Our dataset is specifically designed for the evaluation of
vision based aircraft detection algorithms for fixed-wing aircraft. 
This dataset will allow researchers to evaluate their algorithms
in a systematic and realistic way against existing approaches. Our benchmark data set can also uncover important failure modes of existing methods (such as frame-differencing approaches face with stationary aircraft), and suggest avenues for future research based on a quantitative analysis of algorithm performance or extending recent results learning vision based aircraft detectors \cite{Hussaini} with real data.

\section{Data Collection}\label{sec:dataCollection}

\begin{figure}[!t]
\begin{center}
\includegraphics[scale=0.25]{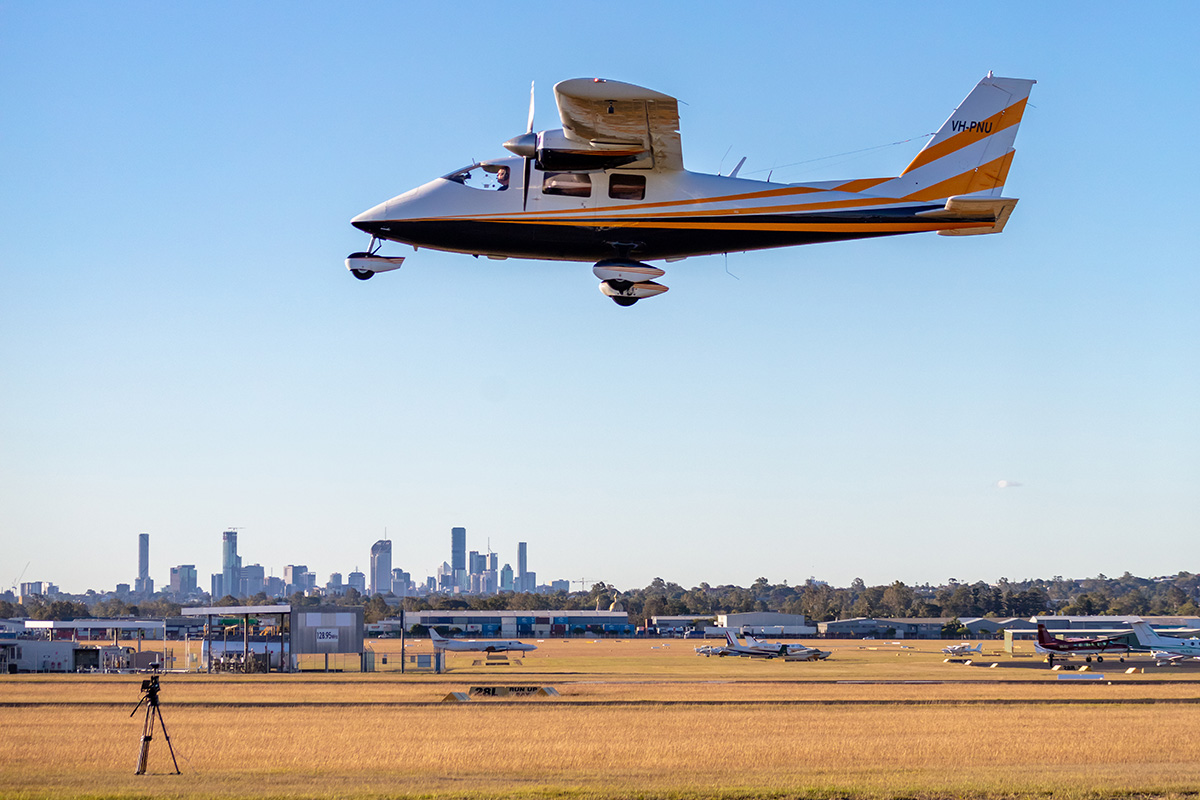}
 \caption{The Partenavia P68 aircraft flying over the camera, as in each image sequence}
\label{flyover}
\end{center}
\end{figure}

Image sequences of two aircraft on true collision course encounter would naturally serve as an ideal test case for this problem. However, due to the obvious risk of flying two aircraft on converging paths this type of data is incredibly scarce and not publicly available. This motivated us to try best capture the properties of a head on collision course (stationary aircraft with respect to the background) by having a single piloted, fixed-wing aircraft converging on a stationary ground camera. This would allow us to capture a realistic example of what would be observed by an aircraft on a collision course scenario and contain sequences depicting a target aircraft emerging from the background. 

Our dataset was collected over two days from 23rd to 24th June, 2020 at Archerfield Airport, Australia, with WGS84 coordinates $27.57^{\circ}S, 153.01^{\circ}E$. 
Our piloted target aircraft was a Partenavia P68 as seen in  Figure \ref{flyover}, and the camera used was a Marshall CV346  set to auto exposure. 

\begin{figure}
\begin{center}
\includegraphics[scale=0.6]{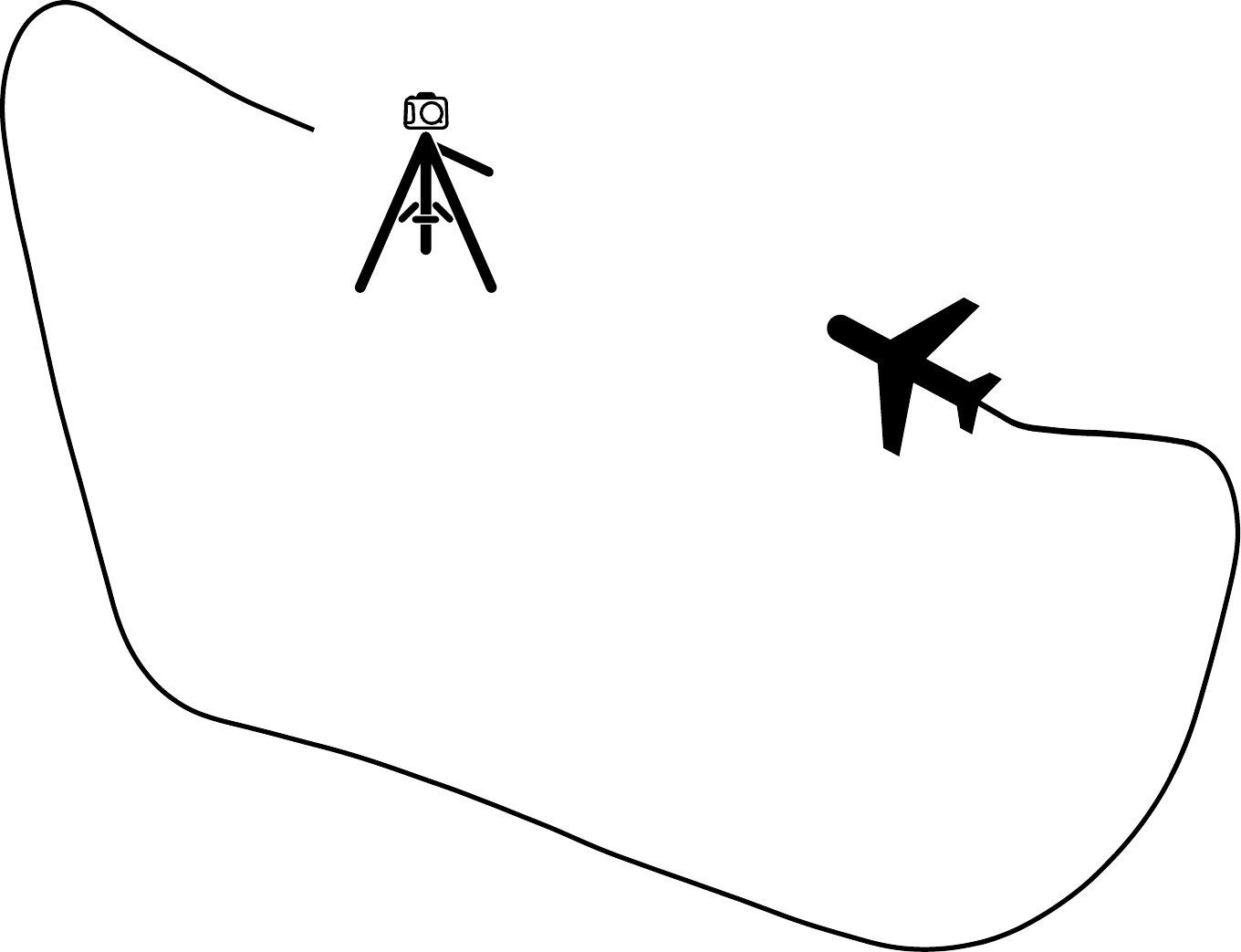}
 \caption{Data collection setup, with real flight track from GPS coordinates}
\label{fig:diag1}
\end{center}
\end{figure}

Flight tests were conducted as follows.  The video data was collected via a camera on the ground positioned under the approach path (extended runway centre line), about 160m East of the runway 28L threshold and mounted on a tripod 1.7m in height.  
Ground crew (who operated the equipment) set up a Mobile Operations Centre (MOC), which is a light truck, about 90m South-East of the camera, next to the taxiway A8 in front of the hangars. A set of cables connected the camera with the MOC. No operator was at the cameras during data capture, except for setup and changes to the configuration. 
Data collection flights consisted of blocks of circuits as seen in Figure \ref{fig:diag1} which outlines the flight path using GPS coordinates.
Each circuit had an extended downwind and final leg varying between  1.6nm (3km) and 3.2nm (6km) from the camera, to ensure that the aircraft was captured emerging from the background.

The encounters were recorded over two days due to rough conditions on the first day. The video was recorded in ProRes 4:2:2 format with UHD resolution a 25 fps frame rate. Recording time was one video per circuit, stopping and restarting the recording when the aircraft passes over the camera. In post processing the videos were exported into still images in TIFF format with resolution of $1280\times 720$. The data has been organised into 15 sequences which we have classified as either \textit{single-aircraft} (sequences S1 through S7) or \textit{multi-aircraft} (sequences M1 through M8). Time and date information for  each sequence is outlined in Table \ref{tbl:info}.

\begin{table*}[!t]
\caption{Summary of Collected Data}
\label{tbl:info}
\renewcommand{\arraystretch}{1.3}
\centering
\begin{tabular}{p{0.3in}|p{0.8in}|p{0.4in}|p{0.9in}|p{0.9in}|p{2.2in}}
\hline
\bfseries Label & \bfseries Collection Date and Local Time  & \bfseries Total Image Frames & \bfseries Target Aircraft Frames &\bfseries  Multi Aircraft Frames & \bfseries  Morphological Characteristics \\
\hline\hline
$S1$ & 23/6/20 (12:11) & 4,423 & 10,671 - 11,728 & N/A &  Flashing bright target aircraft in near-horizon clouds and haze. \\
$S2$  & 23/6/20 (12:19) &  7,733 & 22,601 - 24,548 & N/A & Target aircraft transitions between dim and bright in cloudy sky. \\
$S3$  &  23/6/20 (15:16) &  6,109 &7,696 - 9,544 & N/A & Dim target aircraft in cloudy sky.\\
$S4$  &   24/6/20 (11:20) &  1,196 & 50,400 - 51,587 & N/A & Bright target aircraft in clear sky.  \\
$S5$  &  24/6/20 (11:25) &  1,676 & 58,297 - 59,426 & N/A & Bright target aircraft in clear sky.\\
$S6$  & 24/6/20 (11:31) & 2,053  & 66,406 - 67,513 &  N/A & Bright target aircraft in clear sky. \\
$S7$  & 24/6/20 (11:52) & 5,655 & 98,828 - 10,0067 & N/A & Bright target aircraft in clear sky.\\
\hline
$M1$ & 24/6/20 (11:41) & 2,273 & 81,840 - 83,091 & 81,895 - 83,019 & Bright target aircraft in clear sky, second aircraft flashing in an out of view.\\
$M2$ &  24/6/20 (10:52) & 3,890 & 8,388 - 9,790 & 7,800  - 9,312 &  Bright target aircraft in clear sky, second dim aircraft.\\
$M3$ & 24/6/20 (10:57) & 2,473 & 22,904 - 25,044 & 22,570 - 23,649 \newline 24,430 - 25,044 & Dim target aircraft in clear sky, second dim aircraft with flashing lights, third dim aircraft.\\
$M4$ & 24/6/20 (11:04)  & 2,229 & 25,904 - 27,269 & 24,950 - 25039 \newline 24,950 - 26,319 \newline 26,080 - 27,269  & Bright target aircraft in clear sky, second dim aircraft, third aircraft transitions from dim to bright, fourth dim aircraft. \\
$M5$ &  24/6/20 (11:09) & 5,475 & 34,840 - 35,898 & 30,640 - 32,699 \newline 32,630 - 35,339 \newline 35,580 - 35,898 & Bright target aircraft in clear sky, second aircraft transitions from dim to bright, third aircraft transitions from bright to dim, fourth dim aircraft. \\
$M6$ & 24/6/20 (11:15) & 3,241  & 42,349 - 43,461 & 41,970 - 42959 & Bright target aircraft in clear sky, second bright aircraft.\\
$M7$ & 24/6/20 (11:36) & 5,769 & 74,695 - 75,830 & 70,760 - 72,144 \newline 73,450 - 75,369 &  Bright target aircraft in clear sky, second aircraft transitions from bright to dim, third dim aircraft.\\
$M8$ &  24/6/20 (11:46) & 1,326 & 90,090 - 91,326 & 90,000 - 90,174 &  Bright target aircraft in clear sky, second aircraft both bright and dim components (close to camera).\\
\hline
\end{tabular}
\end{table*}

\section{Vision Based Aircraft Detection Dataset}\label{sec:dataset}
In this section we present our candidate dataset for vision based aircraft detection of small to medium sized, fixed wing aircraft. The complete dataset can be accessed through the website \textit{\href{https://qcr.github.io/dataset/aircraft-collision-course/}{https://qcr.github.io/dataset/aircraft-collision-course/}}. The dataset contains 15 image sequences, ground truth target aircraft labels and  videos of each encounter.

\subsection{Dataset Structure}
Each image sequence contains frames with no aircraft and then captures the emergence of the target aircraft on approach.  
In single-aircraft data there is only one aircraft in the scene on a collision course; the target aircraft. In multi-aircraft data, there is one aircraft on a collision course (the target aircraft), and at least one other aircraft is present in the frame for some of the sequence. In the case of multi-aircraft sequences, only one aircraft is on a collision course with the camera. The image sequences are labeled as a series of frame numbers. The  \textit{Target Aircraft Frames} are given in in Table \ref{tbl:info}. We highlight that the first target aircraft appearance frame was judged by a human oracle as the first frame where the target aircraft is visible however it is potentially in view prior to this and may be detected earlier.

\subsection{Ground Truth Labels}
Each image sequence of images includes a CSV file with ground truth labels. The label is composed of:

\begin{enumerate}
    \item The frame number, 
    \item  the target's approximate centroid location $(u,v)$ in image coordinates, and 
    \item the range of the aircraft in that frame. 
  \end{enumerate}
Our reference frame for image coordinates has its origin at the top-left corner of the image, with $u$ increasing to the right and $v$ increasing down.

The target location labels are a blend of manual and automatic labelling. For each sequence, a number of manual labels were generated. Then the full set of $(u,v)$ locations were linearly interpolated between these manual labels. For improved accuracy, the manual labelling is more dense in: the beginning of the sequence, the end of the sequence, parts of the sequence where the aircraft has more lateral movement.
  
We emphasise that the $(u,v)$ labels are indicative only. Even manual labelling the centroid of the aircraft is subjective, as the target may be poorly delineated from the environment due to atmospheric effects or poor contrast. We recommend the application of the $(u,v)$ labels to verify that a detector is looking in the correct \textit{region} of the image.

\subsection{Dataset Commentary}
This dataset exhibits many useful characteristics for evaluating vision based detection systems including variance in weather, lighting conditions, and morphology of the aircraft.

\begin{figure}
\begin{center}
\includegraphics[scale=2.2,viewport=890 100 940 150, clip=true]{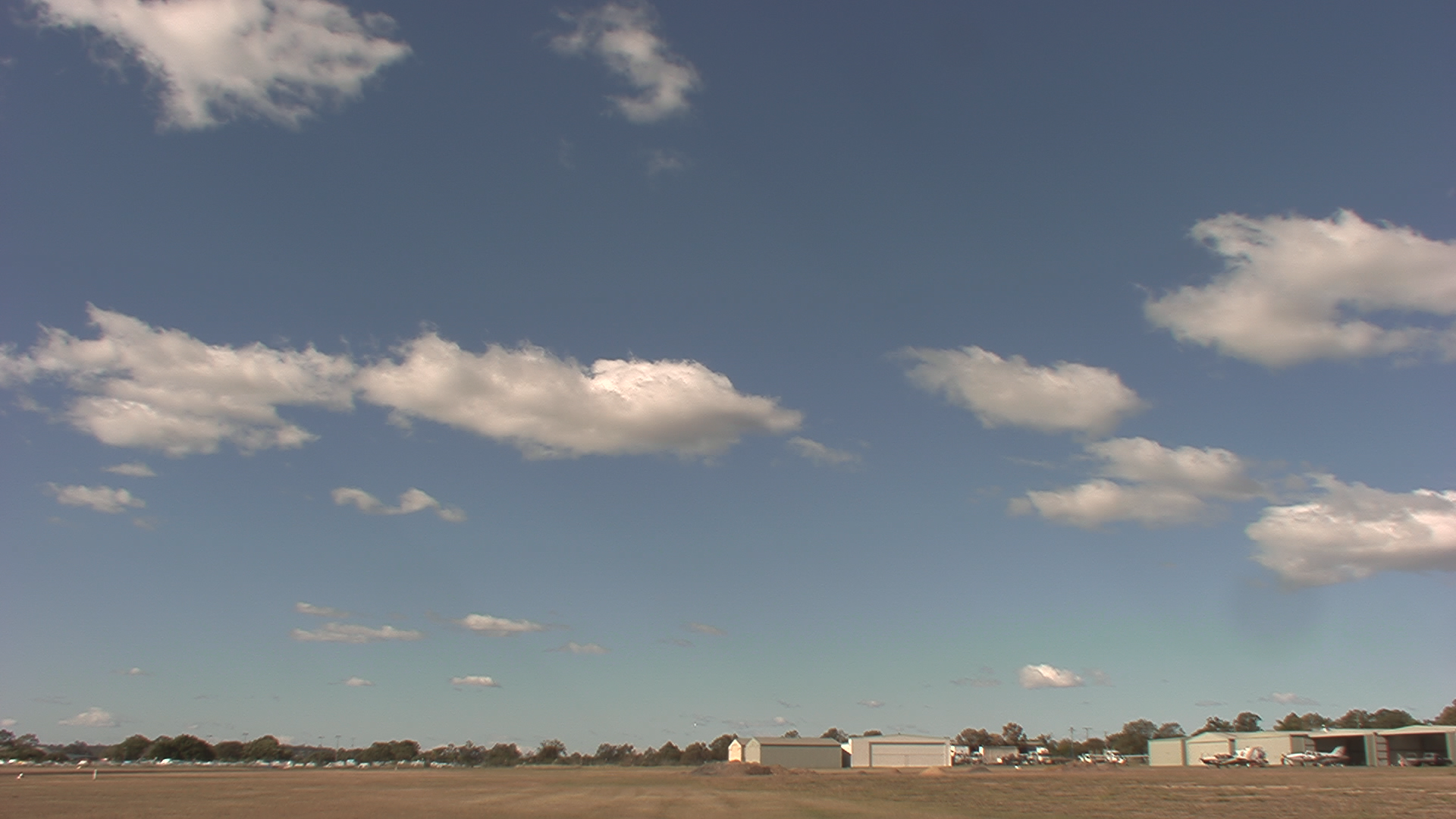}
\includegraphics[scale=2.2,viewport=881 120 931 170, clip=true]{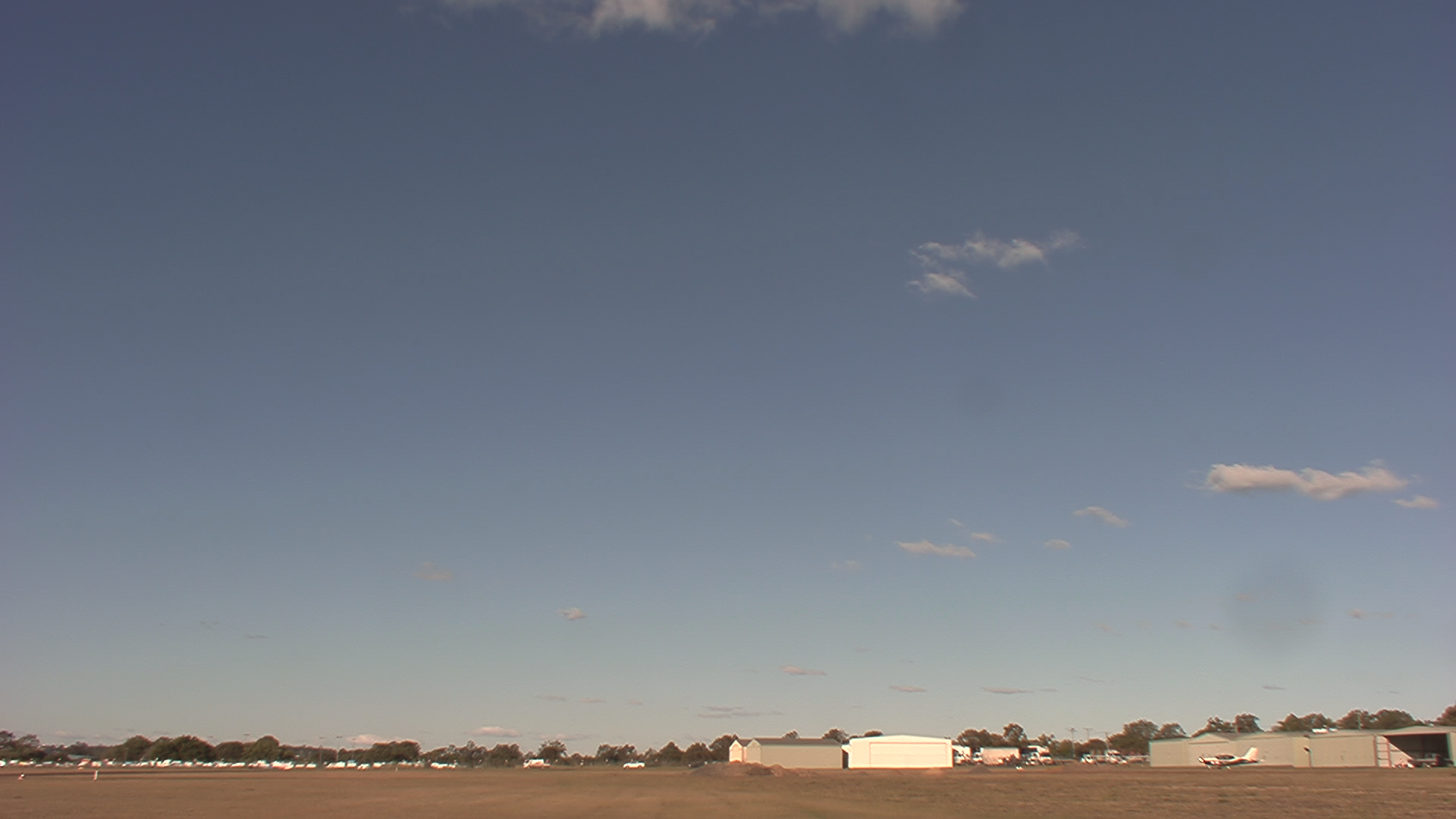}
 \caption{(Left) An example of a bright target aircraft positively contrasting with the background when first coming into view in sequence S1.
 (Right) An example of the aircraft negatively contrasting with the background. In this case, the aircraft resembles adjacent clouds in sequence S3.}
\label{fig:Contrast}
\end{center}
\end{figure}

 We define aircraft morphology as the primitive visual features of the aircraft when it occupies a small region of pixels in the image. For example, the aircraft may either contrast negatively with its environment (appear as a dim region) or positively (appear as a light or bright region, either due to lights on the body of the aircraft or sunlight reflecting off the hull). Our dataset features examples of both as seen in Figure \ref{fig:Contrast}. The aircraft may appear as bright flashing lights, a pale dot, or a dark dot, it may also undergo transitions in morphology as the sequence progresses. We have outline some of the morphological characteristics in Table \ref{tbl:info}.

Weather phenomena such as cloud and haze also pose challenges to aircraft detection. Our dataset contains examples of the aircraft contrasting poorly with the sky due to haze (S1). In some cases, the first visible feature of the aircraft is its landing lights while the aircraft is still obscured by clouds and/ or haze.

\section{Benchmark Performance}\label{sec:benchmark}

In this section we provide a benchmark performance for comparison. 
We exploit the hidden Markov model-based aircraft detector, described in \cite{jamesCST} as ISD-4I. This detector is is composed of four test statistics operating in parallel, where each test statistic represented a probability the the aircraft is in the image frame for a particular motion patch (see \cite{Lai2013} for a detailed explanation on patches). 
 A key point of difference is instead of morphologically processing with the \textit{bottom hat} operation we use the \textit{close minus open} operation described in \cite{lai08} as we have both bright and dim targets.

We highlight that detection range and false alarm performance varies with the choice of the threshold parameters. As done in \cite{jamesCST}, we have identified the lowest thresholds for each algorithm that achieve zero false alarms (ZFAs) for this dataset (the ability to achieve low false alarm rates is consistent with findings in \cite{Lai2011, Lai2013, jamesCST, james2018learning}). In practice, detection thresholds could be adaptively selected on the basis of scene difficulty such as proposed in  \cite{molloy2017thresh}.

\begin{figure}
\begin{center}
\includegraphics[scale=0.62]{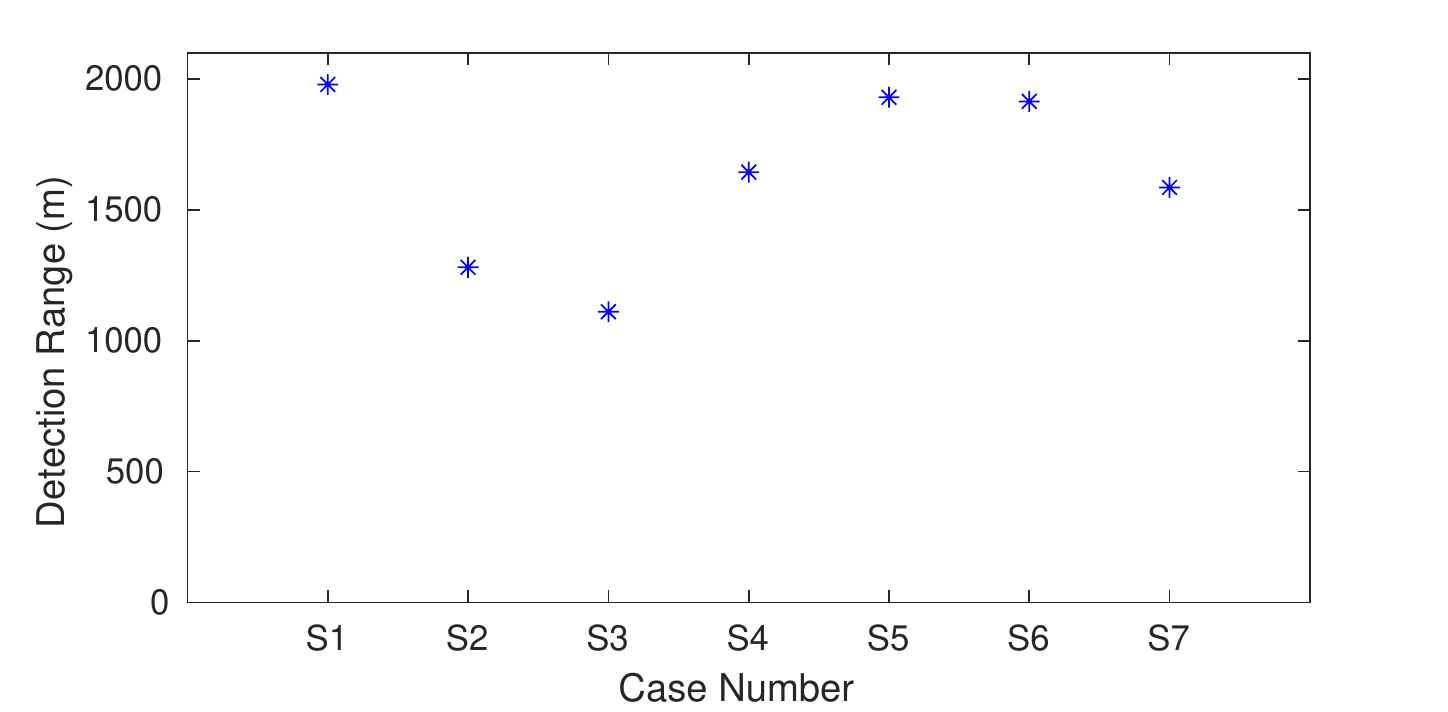}
 \caption{The zero false alarm detection ranges for our benchmark rule ISD-4I on the single target cases (S1-S7).}
\label{fig:detection}
\end{center}
\end{figure}

\begin{figure}
\begin{center}
\includegraphics[scale=0.6]{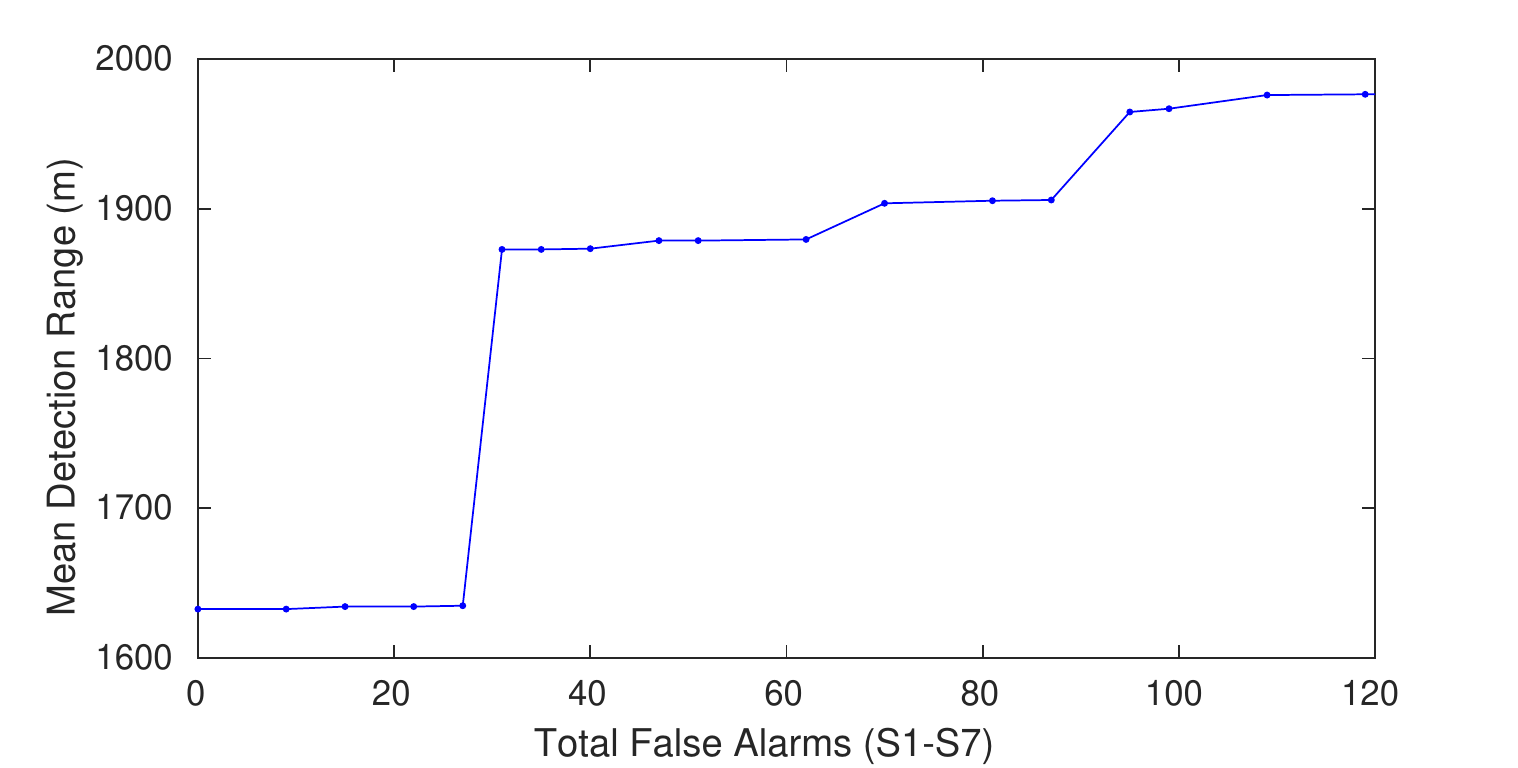}
 \caption{The mean detection ranges (m) and mean false alarms for our benchmark rule ISD-4I on the single target cases (S1-S7).}
\label{fig:SOC}
\end{center}
\end{figure}

The resulting ZFA detection ranges are presented in Figure \ref{fig:detection}.  The mean detection distance and standard error was $1635.4$m and $128$m. To examine the detection range and false alarm performance for different thresholds we also composed system operating characteristic (SOC) curves for the ISD-4I rule presented in Figure \ref{fig:SOC}. We highlight that false alarm was counted for each frame that the test statistic was above the threshold.

\section{Conclusion}\label{sec:conc}
In this paper we presented a dataset for vision based aircraft detection. The dataset consists of 15 uncompressed, high-resolution image sequences containing 55,521 images of a fixed-wing aircraft approaching a stationary, grounded camera. 
Ground truth labels and a performance benchmark are also provided. To our knowledge, this is the first public dataset for studying aircraft on a collision course with the observer.
Our dataset is specifically designed for the evaluation of
vision based aircraft detection algorithms for fixed-wing aircraft. 
This dataset will allow researchers to evaluate their algorithms
in a systematic and realistic way against existing approaches. Our data set can also uncover important failure modes of existing methods (such as frame-differencing approaches face with stationary aircraft), and suggest avenues for future research based on a quantitative analysis of algorithm performance.
\section{Acknowledgements}
This work was enabled by use of the Research Engineering Facility hosted by the Office of Research Infrastructure at QUT, and the use of SmartNet RTK corrections. 
 The authors acknowledge continued support from the Queensland University of Technology (QUT) through the Centre for Robotics.

\bibliographystyle{IEEEtran}
\bibliography{ref}

\end{document}